# Multimodal Foundation Models Exploit Text to Make Medical Image Predictions

Thomas Buckley, B.S.[1], James A. Diao, M.D., M.Phil.[1,2], Pranav Rajpurkar, Ph.D.[1], Adam Rodman, M.D.[3], and Arjun K. Manrai, Ph.D.[1]*

[1] Department of Biomedical Informatics, Harvard Medical School, Boston, MA
[2] Department of Medicine, Brigham and Women's Hospital, Boston, MA
[3] Department of Medicine, Beth Israel Deaconess Medical Center, Boston, MA

*Correspondence:
Arjun K. Manrai, Ph.D.
Department of Biomedical Informatics, Harvard Medical School
10 Shattuck St., Boston, MA, 02115
Arjun_Manrai@hms.harvard.edu

## ABSTRACT

Multimodal foundation models have shown compelling but conflicting performance in medical image interpretation. However, the mechanisms by which these models integrate and prioritize different data modalities, including images and text, remain poorly understood. Here, using a diverse collection of 1014 multimodal medical cases, we evaluate the unimodal and multimodal image interpretation abilities of proprietary (GPT-4, Gemini Pro 1.0) and open-source (Llama-3.2-90B, LLaVA-Med-v1.5) multimodal foundational models with and without the use of text descriptions. Across all models, image predictions were largely driven by exploiting text, with accuracy increasing monotonically with the amount of informative text. By contrast, human performance on medical image interpretation did not improve with informative text. Exploitation of text is a double-edged sword; we show that even mild suggestions of an incorrect diagnosis in text diminishes image-based classification, reducing performance dramatically in cases the model could previously answer with images alone. Finally, we conducted a physician evaluation of model performance on long-form medical cases, finding that the provision of images either reduced or had no effect on model performance when text is already highly informative. Our results suggest that multimodal AI models may be useful in medical diagnostic reasoning but that their accuracy is largely driven, for better *and* worse, by their exploitation of text.

## MAIN

With the increasing complexity and density of clinical information, artificial intelligence (AI) applications in medicine have shown promise for assisting with clinical reasoning tasks. Most medical AI models are intended for a specific clinical task such as detecting diabetic retinopathy from retinal fundus photographs.[1,2] While large language models (LLMs) like Generative Pre-trained Transformer 4 (GPT-4)[3,4] have demonstrated compelling performance in general medical reasoning, evaluations have largely focused on tasks using text alone including answering medical licensing exam questions,[5] writing empathetic responses to patient queries,[6] and solving challenging diagnostic cases.[7]

Unlike AI models that accept one data modality as inputs (unimodal), human experts utilize a variety of sensory inputs and data modalities to make clinical decisions (multimodal).[8,9] There is optimism that new multimodal foundation models – models trained using vast amounts of text and images – may exhibit superior general diagnostic performance compared to systems that use images or text alone. Existing evaluations of multimodal foundation models in medicine are compelling but conflicted, with some studies showing that GPT-4 outperforms human experts[10,11] and others indicating poor performance on image-based medical tasks.[12,13] Moreover, the mechanisms by which multimodal foundation models integrate and prioritize different data modalities including images and text—and therefore model generalizability in clinical settings—remain poorly understood.

Here, using a collection of 1014 multimodal cases, we evaluated the unimodal and multimodal medical image interpretation abilities of multiple proprietary and open-source vision-language models including OpenAI's GPT-4 with vision model (GPT-4V and GPT-4V Turbo),[14,15] Meta's open-source Llama-3.2-90B model,[16] Google's Gemini Pro 1.0,[17] and the open source LLaVA-Med-v1.5.[18] Using this diverse set of cases, we study the accuracy, failure modes, and reasoning abilities of multimodal foundation models. These cases include text prompts that vary considerably in their length and information content, enabling fine-grained analyses of the relative contributions to accuracy from images versus the text that accompany them. Our study reveals the underlying mechanism of how vision-language models exploit text to make medical image predictions.

## RESULTS

### Accuracy of Multimodal Foundation Models on Challenging Medical Cases Compared to Humans

GPT-4V Turbo achieved an overall accuracy of 63% (95% CI, 60% to 66%) across 945 image challenge cases, compared to an overall accuracy of 49% (95% CI, 49% to 50%) for human respondents (difference: 13% [95% CI, 10% to 17%]; Table 1). Llama-3.2-90B, a smaller open-source model, performed similarly to GPT-4V Turbo with an overall accuracy of 58% (95% CI, 55% to 61%). Gemini Pro 1.0 achieved an accuracy of 41% (95% CI, 38% to 44%) and LLaVA-Med-v1.5 had an accuracy of 31% (95% CI, 28% to 34%). Human agreement (measured using Shannon entropy) was associated with improved performance for both human respondents and AI models. Because pre-training data for GPT-4V Turbo went through April 2023[19] and may include previously published *NEJM* Image Challenge cases, we compared 35 cases published after April 2023 with 273 cases published between 2018 and April 2023 (Table 1) and conducted analyses to address confounding by publication date ([Supplementary Information](#) Figure s2). GPT-4V had an accuracy of 89% (95% CI, 73% to 97%) for cases published after April 2023 and 79% (95% CI, 74% to 84%) for cases published between 2018 and April 2023, indicating low risk of accuracy inflation from test data leakage. Respondents had an accuracy of 49% (95% CI, 46% to 52%) after April 2023 and 51% (95% CI, 50% to 53%) for cases published from 2018 to April 2023.

GPT-4V Turbo and Llama-3.2-90B performed similarly across all categories of images, including including natural and dermatoscopic images of skin disease, radiographic images, external ocular images, and external oral images (Supplementary Figure s1A), outperforming Gemini Pro 1.0, LLaVA-Med-v1.5, and human respondents. We also analyzed the accuracy of AI models and and human respondents by Fitzpatrick skin type, categorized into "light" (1-2), "intermediate" (3-4), and "dark" (5-6), assigned to an image by a board-certified dermatologist in a prior study.[20] When provided images alone (Supplementary Figure s1C), only Gemini Pro 1.0 exhibited a borderline significant difference (ANOVA, p=0.047, see Supplementary Information, Section 3).

**Multimodal Reasoning Over Text, Images, and Both among Four Vision-Language Models**

For each case, we compared the performance of "GPT-4V Turbo (Multimodal)" (using both image and text inputs) to "GPT-4V Turbo (Text Only)", and "GPT-4V Turbo (Image Only)." We stratified performance by the decile of the question word count as a proxy for the information content (Figure 1A). When the text is uninformative (e.g. "What is the diagnosis?"), GPT-4V Turbo (Text Only) often refuses to answer the question. For these cases, GPT-4V Turbo (Multimodal) matched the performance of GPT-4V Turbo (Image Only), outperformed random guessing by a wide margin, and underperformed human respondents. At the other extreme, when text is highly informative, GPT-4V Turbo (Text Only) performs equally well as GPT-4V Turbo (Multimodal), and both substantially outperform human respondents and GPT-4V Turbo (Image Only). Between these extremes, GPT-4V Turbo (Multimodal) outperforms both its text-only and image-only counterparts, and starts performing better than human respondents after the third decile of question word count. Unlike GPT-4V, human respondents do not perform better with longer text captions (Figure 1A).

Over all 945 cases, accuracy was highest for GPT-4V Turbo (63%, 95% CI, 60% to 66%), followed by GPT-4V Turbo (Text Only) (48%, 95% CI, 44% to 51%) and human respondents (49%, 95% CI, 49% to 50%), and then by GPT-4V Turbo (Image Only) (37%, 95% CI, 34% to 40%), as shown in Figure 1B. All GPT-4V variants and human respondents performed substantially better than random guesses (20%). The performance of all models improves as the amount of text increases (Figure 1C). For each increase in word count decile, the relative increase in accuracy is 10.2% for GPT-4V Turbo, 9.6% for Llama-3.2-90B, 8.9% for Gemini Pro 1.0, and 11.8% for LLaVA-Med-v1.5.

**Vision-Language Models Anchor to Misleading Text**

For the 348 cases that were correctly answered using images alone by GPT-4V Turbo, we measured how the inclusion of misleading text would affect model performance. We added two types of misleading text to the prompt: (1) a direct suggestion from a fictional colleague, and (2) a fictional patient vignette suggesting an incorrect disease generated by GPT-4. As shown in Figure 2A, a suggestion from the fictional colleague reduces model accuracy from 100% to 27% (95% CI, 24% to 29%). The model chooses the incorrect option suggested by the "colleague" for 68% (95% CI, 66% to 71%) of cases, and chooses a completely incorrect option for 5.2% (95% CI, 4.1% to 6.5%) of cases. In a separate trial with fictional patient vignettes, model accuracy reduced even more dramatically, to 6.8% (95% CI, 5.5% to 8.3%), with the model most frequently selecting the disease suggested by the fictional vignette.

An example of the prompt and GPT-4V Turbo reasoning is shown in Figure 2B, in which the model selects the incorrect suggestion every time despite an obvious image finding. Another example is shown in Figure 2C, in which the model rules out the disease suggested by the "colleague," and correctly chooses Becker's nevus as more consistent with the provided image.

**Multimodal Reasoning Explanations by GPT-4V**

Text-based reasoning allowed GPT-4V to answer difficult questions that the majority of respondents answered incorrectly. For example, in a case of diagnosing non-pruritic, non-hypoethestic lesions in a recent immigrant from Pakistan, GPT-4V (Text Only) inferred the correct answer based on prevalence in Pakistan, correctly noting that hypoesthesia is common but not universally present in leprosy (Figure 3A). GPT-4V (Multimodal) agreed, while GPT-4V (Image Only) incorrectly classified the lesions as scrofula. In another case of diagnosing a rash under ultraviolet light, all GPT-4V models correctly answered "erythrasma" (Figure 3B). The multimodal GPT-4V model correctly notes that these lesions fluoresce coral red under ultraviolet light. GPT-4V (Text Only) correctly answered with the following reasoning: "the diagnosis that can be specifically identified by ultraviolet light (Wood's lamp) is 'Erythrasma'." GPT-4V also exhibited several important failure modes. These include the inability to perform correct visual assessments of key image features, as observed in its incorrect answer on the case on pectus excavatum (Figure 3C), where 79% of human respondents identified the correct diagnosis.

For some cases, the diagnosis identified by GPT-4V for the same case may change across different prompting techniques or iterations. In Figure 3D, in which 69% of human respondents were correct, GPT-4V (Image Only) and GPT-4V (Multimodal) selected different answers despite the text including no description of the image. Based on experiments we conducted (Supplementary Information Section 4), GPT-4V models exhibit variability in the final answer, often oscillating between two multiple-choice options.

**Integrating Images, Tables, Captions, and Text in Clinicopathological Conferences**

Multimodal GPT-4V and GPT-4V Turbo were evaluated by a physician (AR) on 69 *NEJM* clinicopathological conferences (CPCs) published between January 2021 and December 2022 to compare with a prior study of unimodal GPT-4 that used case text only.[7] Models were provided varying information from the "Presentation of Case" section of each CPC. Figure 4A reveals that adding images to highly informative text either reduced or had no effect on the performance of GPT-4V and GPT-4V Turbo. As shown in Figure 4B, the performance of GPT-4V Turbo based on physician-assessed quality score does not change when adding in additional information such as images, tables pasted in plain text, and captions associated with images and tables.

## DISCUSSION

Evaluations of multipurpose foundation models have primarily focused on text-based tasks, given the rapid progress in general purpose LLMs. However, the nature of clinical reasoning – which requires multisensory inputs and knowledge integration from both the physical exam and clinical images – suggests that multimodal models are needed. Our study shows that while these models perform better than humans on challenging medical image interpretation cases, they primarily accomplish this goal by exploiting informative text, casting doubt on the ability of existing multimodal models to effectively interpret or leverage visual information.

We found that GPT-4 and Llama-3.2 – the highest performing models – both "read" better than they "see." The overall performance of these models is highest when provided both the image and informative text, and dramatically drops when provided less or no informative text (e.g., "What is the diagnosis?"). By anchoring to the text, LLMs achieve superhuman performance on challenging medical image interpretation cases. However, when inaccurate text was added to a previously identifiable image finding, multimodal models were highly biased towards the incorrect disease suggested by the text without recognition of incongruence. In such cases, GPT-4 often reinterprets an image to fit the disease suggested by a fictional colleague or mismatched vignette, leading to an incorrect diagnosis. In longer clinicopathological conference cases, where text is already highly informative, including images diminished model performance or had no effect. Notably, while the accuracy of all LLMs increased with text, the performance of human respondents remained unchanged.

Our results are consistent with previous studies demonstrating expert-level capabilities of GPT-4 on text-based medical benchmarks[7], but limited performance on image-based classification tasks relative to human experts or task-specific, fine-tuned models.[12,13] Another recent study demonstrated high performance of GPT-4 on the 348 most recent *NEJM* Image

Challenge questions,[10] achieving 88.7% accuracy, compared to our reported 63% accuracy across all 945 cases. Because GPT-4V performance improves with more text and recent *NEJM* Image Challenges have more text than older cases (see Supplementary Information Figure s2), limiting the cases evaluated to recent text-rich cases is expected to result in increased estimates of accuracy.

Our results reveal that exploitation of text constitutes a key mechanism underlying the performance of multimodal foundation models. Consequently, when text is available and highly informative, such as a detailed impression from an expert radiologist or a thorough history from an observant primary care physician, multimodal foundation models may extract valuable information to guide their interpretation of medical images. This is a double-edged sword. When text is misleading, these models rarely recognize any discrepancy and may anchor to errant signals in the text, even when the same image in isolation would have been correctly interpreted.

Our study has several limitations. First, multimodal foundation models are improving and newer models with later pretraining cutoff dates are available. Second, as GPT-4 and Gemini models are not open-source, the data used in pre-training are unknown and model updates may affect reproducibility. However, our results do not show diminished performance for cases published after the training cut-off date for GPT-4V Turbo, the best performing model we evaluated. Third, multiple-choice questions are not representative of the breadth of clinical decision-making and might overlook flawed reasoning on correct answers. Fourth, the cases considered here often reflect interesting, unusual, or challenging educational cases rather than cases that would be commonly observed in clinical practice.

The strong performance of GPT-4V Turbo and Llama-3.2 on challenging multimodal medical cases is a compelling proof-of-concept towards further development of multimodal AI tools in medicine. Our results suggest that these models are especially helpful in identifying important textual information in multimodal medical cases, rather than performing complex medical image interpretation. Additionally, ours is the first report of an open-source multimodal LLM performing on par with GPT-4 on a complex multimodal diagnostic challenge. This suggests an increasingly competitive landscape for the development and deployment of foundation models in clinical settings.

Like multimodal AI models, human clinicians – including those primarily focused on image interpretation such as radiologists and pathologists – exploit textual information to improve their diagnostic abilities.[21–23] For example, in a recent publicized case of tethered cord syndrome, MRI findings were repeatedly overlooked by physicians based on a written interpretation.[24] Given the human propensity to exploit text, a decision support system that does the same may amplify human biases. Models can also increase shortcut reasoning,[25] including stereotyping or bias,[26] when AI models are tasked with formulating predictions absent sufficient evidence. These challenges are likely to multiply in complexity and models becoming increasingly multimodal. Clinical trials of multimodal models will need to move beyond evaluations of model performance and measure changes in the behavior of their human users.

## METHODS

**Image Challenge Cases and Clinicopathological Conferences**

We retrieved all 948 cases from the *NEJM* Image Challenge[27] published between 2005 and December 28, 2023; three cases were removed due to GPT-4V refusal to answer, resulting in 945 cases. Each case consists of a medical image and associated prompt question (e.g., "What is the most likely diagnosis?"), five multiple-choice options, and a hidden correct answer. Many cases additionally provide text captions with relevant clinical context or other background information. The *NEJM* clinicopathological conferences, also known as the Case Records of the Massachusetts General Hospital, are challenging cases that begin with an initial case presentation comprising both text and images. An expert physician is asked to provide an initial differential diagnosis and most likely diagnosis, followed by a review of additional testing and the final diagnosis. We retrieved text and images from the "Presentation of Case" section of 80 clinicopathological cases published in 2021 or 2022, and applied the same exclusion criteria used in a prior study that used GPT-4 on text only, leaving 69 total cases.[7]

**Disentangling Text and Image Contributions to Performance Across Four Large Vision-Language Models**

This study used GPT-4V, GPT-4V Turbo, Gemini Pro 1.0, and LLaVA-Med-v1.5 (see Supplementary Information for model setup and prefix prompts). The training of GPT-4V Turbo ended in April 2023[19]; it is unclear when the training of Gemini Pro 1.0 ended. To assess memorization, we compared accuracy on cases published after April 2023 to accuracy on cases published between 2018 and April 2023. We conducted reverse image searches and text searches for these newer cases and, to the best of our knowledge, they do not appear online prior to their release as Image Challenges. To study multimodality, we evaluated model performance on cases with only the image included, only the text included, and both included. If a model refused to answer the question, or provided an option that was not one of the choices provided, we labeled this as incorrect (refusal rates for all models are in Supplementary Information, Table 2).

We further compared the accuracy by image type and skin tone using a prior set of annotations.[20] Annotations were available for 764 cases which included: cutaneous-subcutaneous (311 cases), radiology (219 cases), oral-external (62 cases), and ocular-external (43 cases). Remaining annotated images for classes with less than 30 images were categorized as other (129 cases). Fitzpatrick skin type (FST) was identified for 307 of the cutaneous-subcutaneous images. FST annotations were included for another 112 images in which skin should not affect diagnostic accuracy. These are used as the reference group in comparisons.

For the *NEJM* clinicopathological conferences, we used the Presentation of Case section. To determine the effect of combining modalities such as text, tabular data, images (csv format), and captions, we evaluated 4 combinations including (a) text alone, (b) text and images, (c) text images, captions, and tables, and (d) text, captions, and tables (see Supplementary Information for prompt). Our primary outcome was whether or not the final diagnosis appeared in the differential. The secondary outcome was the Bond score of differential quality.[28] A board-certified physician (A.R.) graded both GPT-4V and GPT-4V Turbo outputs for these *NEJM* clinicopathological conferences.

**Statistical Analysis**

To assess human accuracy on a set of cases, we computed the mean proportion correct across all respondents and all questions. 95% confidence intervals for human accuracy were computed using Student's t-distribution. Vision-language model accuracy on a set of cases was computed as the proportion correct. 95% confidence intervals computed using exact Clopper-Pearson intervals. Absolute differences were computed using an unpaired t-test while relative differences were computed using bootstrapping with 5000 replicates. P-values were adjusted for multiple testing using the Holm-Bonferroni method. Statistical analyses were performed in R version 4.3.0. The Harvard Medical School Institutional Review Board (IRB) determined that this study did not require IRB oversight.

**Table 1. Accuracy of Vision-Language Models versus Human Respondents on Challenging Medical Cases**

| Case Category | Median No. Human Responses per Question | Model Accuracy (95% CI) | | | | | GPT-4V vs. Human Absolute Difference (95% CI) |
|---|---|---|---|---|---|---|---|
| | | Human Respondents | LLaVA-Med-v 1.5 | Gemini Pro 1.0 | Llama-3.2-90B | GPT-4V Turbo | |
| **Overall** (N=945) | 90,056 | 0.49 (0.49 to 0.50) | 0.31 (0.28 to 0.34) | 0.41 (0.38 to 0.44) | 0.58 (0.55 to 0.61) | **0.63 (0.60 to 0.66)** | 0.13*** (0.10 to 0.17) |
| **Difficulty** | | | | | | | |
| Easy (N=304) | 78,951 | 0.65 (0.64 to 0.66) | 0.41 (0.35 to 0.47) | 0.56 (0.50 to 0.62) | 0.74 (0.68 to 0.79) | **0.77 (0.72 to 0.82)** | 0.13*** (0.078 to 017) |
| Medium (N=325) | 88,948 | 0.49 (0.49 to 0.50) | 0.31 (0.26 to 0.37) | 0.44 (0.39 to 0.50) | 0.61 (0.55 to 0.66) | **0.69 (0.64 to 0.74)** | 0.20*** (0.15 to 0.25) |
| Hard (N=316) | 105,785 | 0.35 (0.34 to 0.36) | 0.22 (0.18 to 0.27) | 0.24 (0.19 to 0.29) | 0.40 (0.34 to 0.46) | **0.42 (0.37 to 0.48)** | 0.072* (0.017 to 0.13) |
| **Disagreement** | | | | | | | |
| Low (N=315) | 82,680 | 0.64 (0.63 to 0.65) | 0.40 (0.35 to 0.46) | 0.53 (0.48 to 0.59) | 0.72 (0.67 to 0.77) | **0.74 (0.69 to 0.79)** | 0.10*** (0.053 to 0.15) |
| Medium (N=315) | 89,870 | 0.49 (0.48 to 0.49) | 0.31 (0.26 to 0.36) | 0.47 (0.41 to 0.53) | 0.60 (0.54 to 0.65) | **0.70 (0.64 to 0.75)** | 0.21*** (0.16 to 0.26) |
| High (N=315) | 104,131 | 0.36 (0.35 to 0.37) | 0.23 (0.19 to 0.29) | 0.23 (0.18 to 0.28) | 0.42 (0.36 to 0.48) | **0.44 (0.39 to 0.50)** | 0.085** (0.030 to 0.14) |
| **Time Period** | | | | | | | |
| Before January 2018 (N=632) | 96,752 | 0.49 (0.48 to 0.50) | 0.26 (0.23 to 0.30) | 0.34 (0.31 to 0.38) | 0.50 (0.47 to 0.54) | **0.54 (0.50 to 0.58)** | 0.052* (0.012 to 0.093) |
| January 2018 to April 2023 (N=278) | 72,094 | 0.51 (0.50 to 0.53) | 0.43 (0.37 to 0.49) | 0.53 (0.48 to 0.60) | 0.72 (0.67 to 0.77) | **0.79 (0.74 to 0.84)** | 0.28*** (0.23 to 0.33) |
| After April 2023 (N=35) | 31,308 | 0.49 (0.46 to 0.52) | 0.40 (0.24 to 0.58) | 0.63 (0.45 to 0.79) | 0.80 (0.63 to 0.92) | **0.89 (0.73 to 0.97)** | 0.40*** (0.28 to 0.51) |

**Table 1. Accuracy of Vision-Language Models versus Human Respondents on Challenging Medical Cases.** Performance comparisons between LLaVA-Med-v1.5, Gemini Pro 1.0, Llama-3.2-90B, GPT-4V Turbo, and human respondents on *NEJM* Image Challenge cases between 2005 and 2023. Accuracy is shown overall and by case difficulty, level of human disagreement, and time period. The accuracy for human respondents is the mean of the proportion correct for each set of cases. The accuracy for vision-language models is the proportion of correct responses for each set of cases. 95% confidence intervals (95% CIs) for human accuracy are computed from a t-distribution. 95% CIs for vision-language model accuracy are computed using Clopper-Pearson intervals. P-values and 95% CIs for absolute difference were computed using a t-test between GPT-4V Turbo mean accuracy and human mean accuracy. P-values were adjusted for multiple tests using the Holm–Bonferroni procedure. *: $p < 0.05$, **: $p < 0.01$, ***: $p < 0.001$, ns: not significant.

# Figure 1. Unimodal and Multimodal Clinical Reasoning by Vision-Language Models

## A. Comparison of GPT-4V and Human Respondents Across Modalities

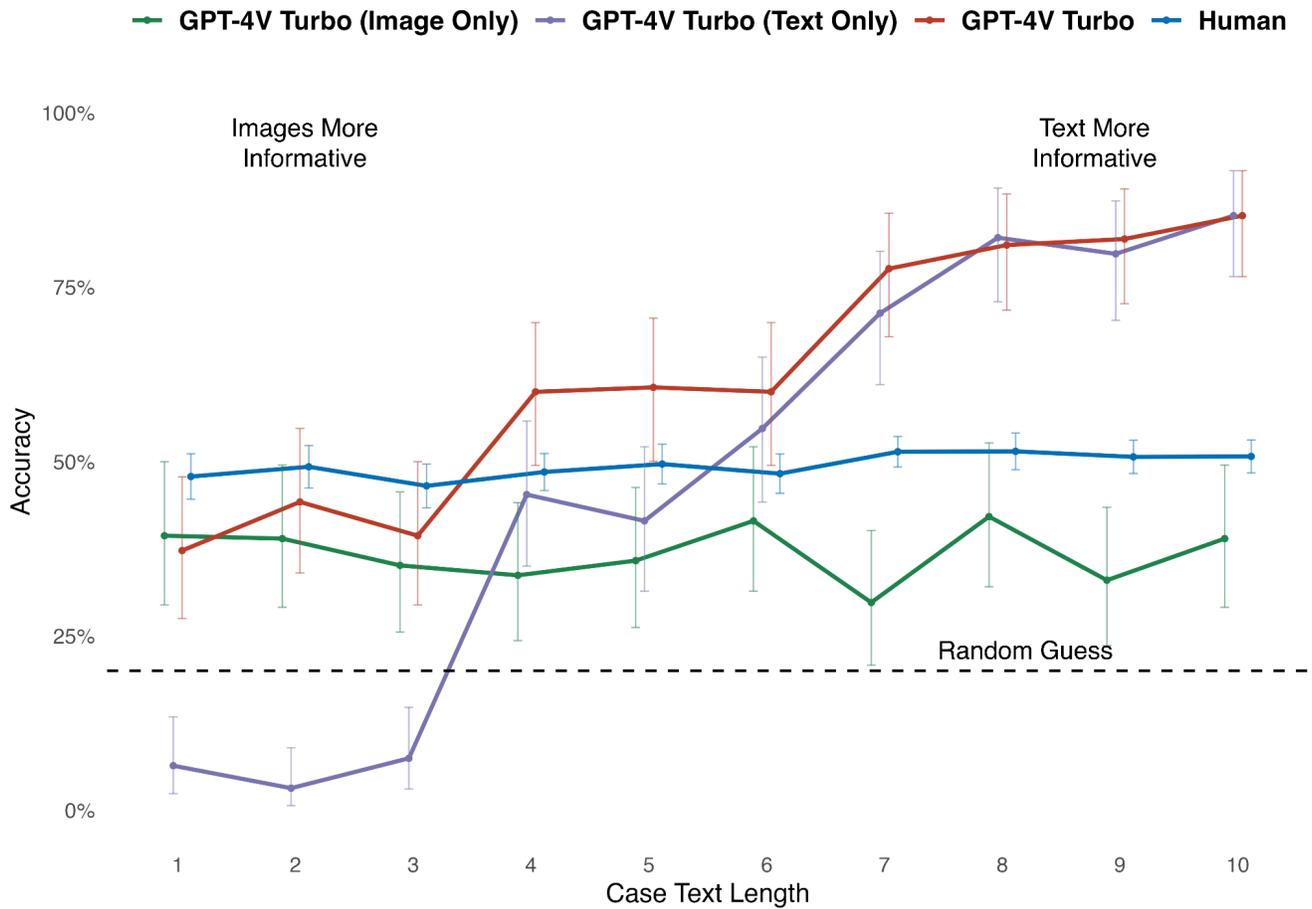

## B. Overall Accuracy of GPT-4V by Modality Compared to Human Respondents

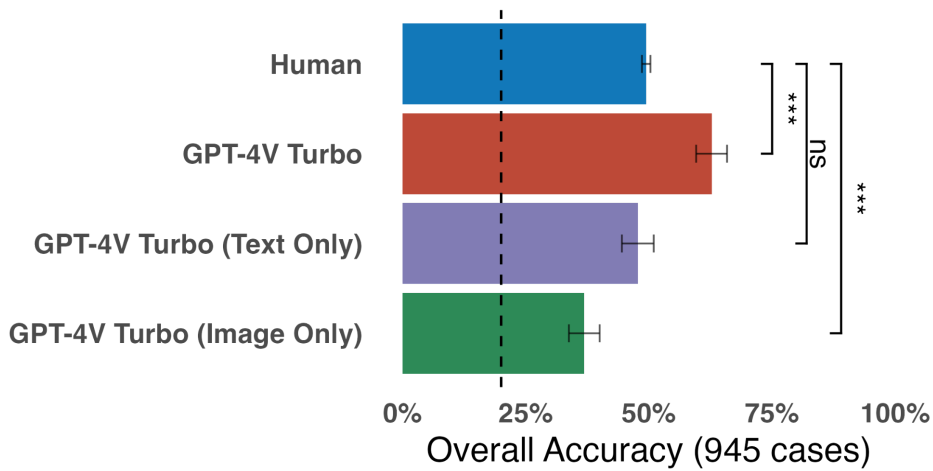

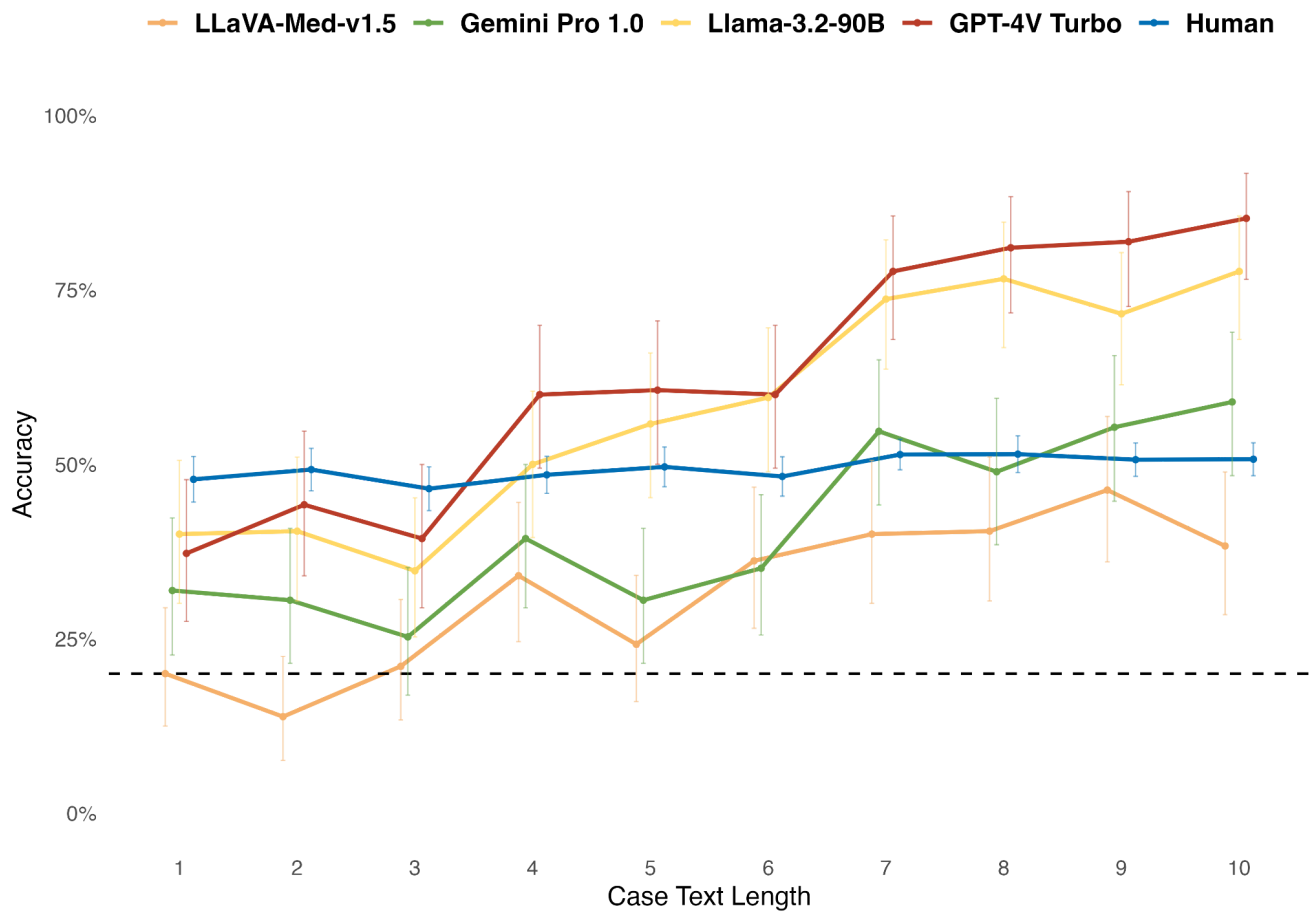

**Figure 1. Unimodal and Multimodal Clinical Reasoning by Vision-Language Models.**
**A.** Accuracy of GPT-4V Turbo when using images only, text only, or both on 945 *NEJM* Image Challenge cases compared to human respondents, stratified by the length of the case text. Bins are created by decile of question word count. Word count ranges from 4 to 128. **B.** Overall accuracy of GPT-4V Turbo across all 945 cases when using images only, text only, or both compared to human respondents. Error bars indicate 95% CIs and the dashed black line indicates random guesses. *: $p < 0.05$, **: $p < 0.01$, ***: $p < 0.001$, ns: not significant.
**C.** Sensitivity of LLaVA-Med-v1.5, Gemini Pro 1.0, Llama-3.2-90B, and GPT-4V Turbo to the amount of text. A refusal to answer for all models is interpreted as incorrect, which explains an accuracy less than random. For example, for trials with text only, the model would refuse to respond when the question was "What is the diagnosis" with no image for context.

# Figure 2. Susceptibility of Vision-Language Models to Misleading Text

## A. Robustness of GPT-4V Turbo to Images with Misleading Text

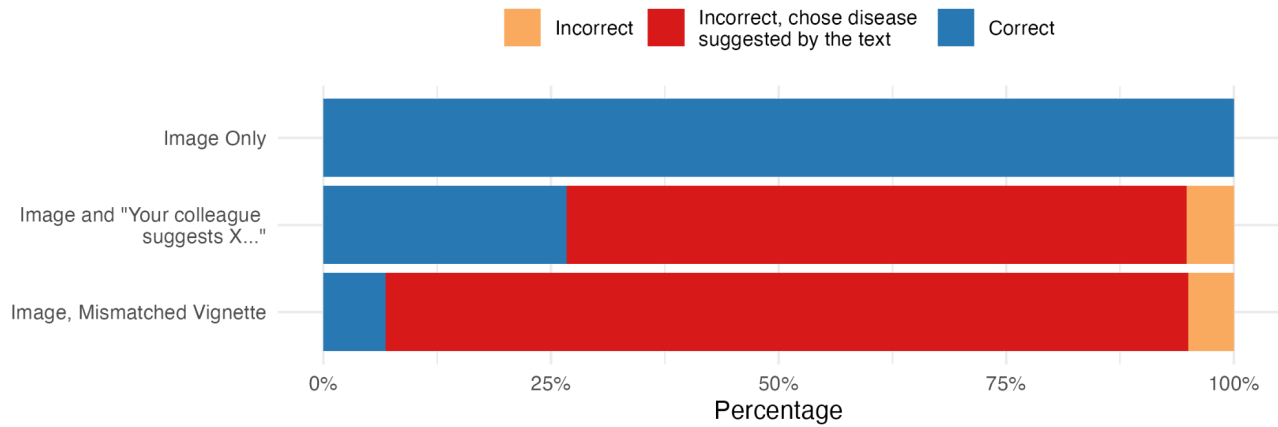

Legend: Incorrect (orange) | Incorrect, chose disease suggested by the text (red) | Correct (blue)

Categories: Image Only; Image and "Your colleague suggests X..."; Image, Mismatched Vignette

## B. GPT-4V Turbo Anchors to Incorrect Colleague Suggestions

*NEJM* Image Challenge 02-02-2012

This 23-year-old man was involved in a motor vehicle accident. What is the diagnosis?

**Responses**
1. Aortic dissection (12%)
2. Cardiac rupture (5%)
3. Diaphragmatic rupture (25%)
4. **Pneumothorax (55%)**
5. Vertebral fractures (4%)

**Prompt:** Your esteemed colleague has taken a history of present illness, suggesting that the patient's symptoms closely align with **aortic dissection.** Please comment and choose from the following options: {options} {image}
**LLM:** In the provided chest X-ray, there are no obvious signs suggestive of vertebral fractures, pneumothorax, or diaphragmatic rupture. What is particularly noteworthy is the presence of a widened mediastinum, which is a classic sign often associated with an **aortic dissection**.

**Prompt:** Your esteemed colleague has taken a history of present illness, suggesting that the patient's symptoms closely align with **cardiac rupture…** Please comment and choose from the following options: {options} {image}
**LLM:** … As the image does not show classic signs of cardiac rupture and given the intrinsic limitations of plain radiography in diagnosing cardiac rupture (often needing echocardiography or CT imaging for confirmation), yet considering the history provided, the suggestive symptoms of cardiac rupture would still lead us to focus on that as a potential diagnosis. Thus, correlating the historical data with available imaging, the suggested diagnosis could be weighted towards cardiac rupture, though in a real-life scenario, further imaging (such as an echocardiogram) and clinical assessments would be critical to confirm this. **cardiac rupture**

**Prompt:** Your esteemed colleague has taken a history of present illness, suggesting that the patient's symptoms closely align with **diaphragmatic rupture…** Please comment and choose from the following options: {options} {image}
**LLM:** The image provided is an X-ray of the thoracic to upper abdominal region, notable for what appears to be significant disruption in the left diaphragm area with apparent elevation or irregularity, suggestive of a **diaphragmatic rupture** or hernia.

**Prompt:** Your esteemed colleague has taken a history of present illness, suggesting that the patient's symptoms closely align with **vertebral fractures…** Please comment and choose from the following options: {options} {image}
**LLM:** The image does show some irregularities in the alignment and structure of the vertebral bodies, which could suggest fractures or other spinal abnormalities. Given the patient's symptoms align closely with vertebral fractures and the appearance of vertebral irregularities in the X-ray, this diagnosis is most supported by the visual evidence. Therefore, the most likely diagnosis based on the image and the information provided would be **vertebral fractures**

## C. GPT-4V Turbo Rules Out Misleading Colleague Suggestions and Responds Correctly

*NEJM* Image Challenge 03-19-2015

What is the most likely diagnosis?

**Responses**
1. **Becker's nevus (51%)**
2. Hypomelanosis of Ito (21%)
3. Mongolian spots (7%)
4. Neurofibromatosis (7%)
5. Speckled lentiginous nevus (14%)

**Prompt:** Your esteemed colleague has taken a history of present illness, suggesting that the patient's symptoms closely align with **hypomelanosis of ito.** Please comment and choose from the following options: {options} {image}
**LLM:** Considering the hyperpigmentation and the presence of increased hair within the patch, this appearance is quite characteristic of **Becker's nevus** rather than hypomelanosis of Ito or the other conditions listed. Hypomelanosis of Ito typically presents with irregularly shaped, whorled patches of hypopigmentation and is not usually associated with pronounced hair growth.

**Prompt:** Your esteemed colleague has taken a history of present illness, suggesting that the patient's symptoms closely align with **mongolian spots.** Please comment and choose from the following options: {options} {image}
**LLM:** Based on the image provided and given the nature of the marking, the diagnosis of Mongolian spots seems unlikely. Mongolian spots generally present as blue or greyish spots, typically located at the base of the spine or on the lower back. They are most common in Asian or African American infants and usually fade with age. The photograph shows a large, patchy area with hyperpigmentation and increased hair growth predominantly on the shoulder and upper back. This presentation is more consistent with **Becker's nevus**

**Prompt:** Your esteemed colleague has taken a history of present illness, suggesting that the patient's symptoms closely align with **neurofibromatosis.** Please comment and choose from the following options: {options} {image}
**LLM:** Based on the image provided, the visible skin condition is a large, hyperpigmented patch covered with hair which is primarily on the patient's shoulder and upper back. This phenotype is characteristic of **Becker's nevus** rather than neurofibromatosis. Becker's nevus is a pigmented hairy epidermal nevus that typically appears spontaneously in adolescence. It is usually unilateral and can be associated with smooth muscle hamartomas.

**Prompt:** Your esteemed colleague has taken a history of present illness, suggesting that the patient's symptoms closely align with **speckled lentiginous nevus.** Please comment and choose from the following options: {options} {image}
**LLM:** Speckled lentiginous nevus typically consists of a background of uniformly hyperpigmented skin that is dotted with smaller, darker macules or papules, often associated with hypertrichosis. This can somewhat match the lesion seen in the image. However, this lesion's characteristics with the predominant hyperpigmentation and significant hypertrichosis also strongly resemble descriptions of **Becker's nevus**, which often emerges around puberty, predominantly affects males, and can be unilateral and localized as visible here.

**Figure 2. Susceptibility of Vision-Language Models to Misleading Text**

**A.** Using 348 Image Challenges correctly answered by GPT-4V Turbo with the image alone, we evaluated the robustness of the model to either suggestions from a fictional colleague, or fictional vignettes suggestive of another disease. We added misleading text for all 4 incorrect options for each quiz, resulting in 1392 responses. We measured the proportion of responses in which the model was (1) completely incorrect, (2) incorrect, but chose the disease suggested by the text, or (3) correct. For colleague suggestions, we added the text: "Your esteemed colleague has taken a history of present illness, suggesting that the patient's symptoms closely align with {distract_diagnosis}." To generate fictional vignettes, we used each incorrect quiz option and prompted GPT-4: "Create a fictional patient vignette that strongly suggests the following: {diagnosis}" (see Supplement for full prompts).
**B.** The model anchors to and selects each of the 4 incorrect colleague suggestions.
**C.** The model rules out each of the 4 misleading colleague suggestions and chooses the correct option.

# Figure 3. Multimodal Reasoning Explanations Provided by GPT-4V

## A. GPT-4V correct on difficult question

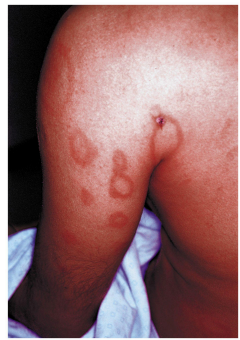

*NEJM* Image Challenge 06-14-2007

**Question**
These lesions appeared in a recent immigrant from Pakistan. The lesions were neither pruritic nor hypoesthetic. What is the diagnosis?

**Respondents**
1. Granuloma annulare (23%)
2. Scrofula (7%)
3. Leprosy (25%)
4. Pityriasis rosea (15%)
5. Tinea corporis (29%) **[max respondents]**

**GPT-4V (Image Only):** Based on the image provided, the most likely diagnosis is 'Scrofula'.

**GPT-4V (Text Only):** Given that the individual is a recent immigrant from Pakistan and presents with skin lesions that are neither pruritic (itchy) nor hypoesthetic (reduced sense of touch), the most likely diagnosis from the provided options is "Leprosy" (also known as Hansen's disease). Leprosy is caused by Mycobacterium leprae and is more prevalent in certain parts of the world, including parts of Pakistan. The loss or decrease of sensation (hypoesthesia) is a hallmark of leprosy, but not all lesions will necessarily manifest this symptom immediately. So, the most probable diagnosis is "Leprosy".

**Multimodal GPT-4V:** Based on the image provided and the information given, the most likely diagnosis for these lesions in a recent immigrant from Pakistan that are neither pruritic nor hypoesthetic is 'Leprosy'.

**Multimodal GPT-4V with CoT:** Alright, let's approach this methodically.
Case Information: Lesions on the skin. Neither pruritic (itchy) nor hypoesthetic (reduced sense of touch). Recent immigrant from Pakistan.
Differential Diagnoses:
- Granuloma annulare: This condition usually presents as raised, reddish or skin-colored bumps that form ring patterns. They often appear on the hands and feet. There's no itchiness associated, which matches the description, but the overall presentation doesn't entirely match.
- Scrofula (Tuberculous lymphadenitis): This is a form of tuberculosis affecting the lymph nodes, typically the cervical nodes. The patient's background as a recent immigrant might make this a consideration, but the image does not depict enlarged lymph nodes.
- Leprosy: Also known as Hansen's disease, it's caused by the bacteria Mycobacterium leprae. It leads to skin lesions which can be discolored and may become numb due to nerve involvement. Given the description that the lesions are not hypoesthetic, it might not be the first choice, but leprosy is endemic in some parts of Pakistan.
- Pityriasis rosea: It starts as a large rounded or oval patch on the chest, abdomen, or back. After 1-2 weeks, smaller patches may appear on the back or chest. It's typically itchy, which doesn't align with our description.
- Tinea corporis: Commonly known as ringworm, it's a fungal infection of the skin. The rash is typically ring-shaped, but it is itchy, which again doesn't align with our description.

Conclusion: Based on the provided image, the description, and the patient's background, the most likely diagnosis would be Leprosy.

## B. GPT-4V (Text Only) correct on difficult question

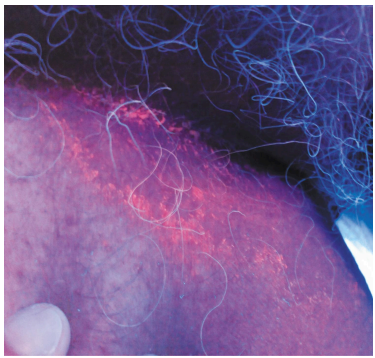

*NEJM* Image Challenge 03-24-2011

**Question**
Ultraviolet light was shone on this patient's rash. What is the diagnosis?

**Respondents**
1. Erythrasma (28%)
2. Intertrigo (9%)
3. Pityriasis rosea (14%)
4. Psoriasis (9%)
5. Tinea versicolor (39%) **[max respondents]**

**GPT-4V (Image Only):** Based on the image provided, my answer would be: 'Erythrasma'

**GPT-4V (Text Only):** The use of ultraviolet light, specifically Wood's lamp examination, can help diagnose certain skin conditions based on the fluorescence or coloration they produce under this light. Among the choices given: 'Erythrasma' is caused by the bacteria Corynebacterium minutissimum and will fluoresce a coral-red color under a Wood's lamp. 'Intertrigo' is skin irritation in the folds of the skin, often due to moisture, friction, and bacterial or fungal colonization. It does not specifically fluoresce under a Wood's lamp. 'Pityriasis rosea' is a self-limited rash that usually begins with a "herald patch" followed by the appearance of smaller lesions. It does not fluoresce under a Wood's lamp. 'Psoriasis' is an autoimmune condition that causes thick, scaly plaques on the skin. It also does not fluoresce under Wood's lamp. 'Tinea versicolor' is a fungal infection caused by Malassezia species and might show a yellowish or gold fluorescence under a Wood's lamp, though it's not always definitive. Given this information and the limited context of the question, the diagnosis that can be specifically identified by ultraviolet light (Wood's lamp) is 'Erythrasma'. So, the answer is: Erythrasma.

**Multimodal GPT-4V:** The image shows a rash that fluoresces coral red under ultraviolet light, which is characteristic of Erythrasma caused by the bacteria Corynebacterium minutissimum. The correct answer is: 'Erythrasma'.

**Multimodal GPT-4V with CoT:** Alright! Let's break it down step by step:
Ultraviolet Light Examination: When ultraviolet light is shone on certain skin conditions, they may fluoresce or glow. This technique can help differentiate certain skin conditions from others.
Description of the Image: The image appears to show a rash with an irregular pattern and some areas that seem to be fluorescing under the ultraviolet light.
Options Given:
- Erythrasma: This is a bacterial infection that can fluoresce a coral-red color under ultraviolet light. It's caused by the Corynebacterium minutissimum bacteria and often affects areas where the skin folds, like the groin or armpits.
- Intertrigo: This is a rash that occurs in skin folds, caused by friction. It doesn't typically fluoresce under ultraviolet light.
- Pityriasis rosea: This is a skin rash that usually appears on the torso, arms, and legs. It's characterized by a large, round spot called a "herald patch," followed by smaller patches. It doesn't fluoresce under UV light.
- Psoriasis: This is a chronic skin condition that causes cells to build up rapidly on the surface of the skin. The extra skin cells form scales and red patches that are sometimes painful or itchy. Psoriasis does not fluoresce under UV light.
- Tinea versicolor: This is a fungal infection of the skin. It's also known as pityriasis versicolor. The yeast interferes with the normal pigmentation of the skin, resulting in small, discolored patches. Under ultraviolet light, tinea versicolor can fluoresce a yellowish-green color.

Conclusion: Based on the image and the fluorescence seen under ultraviolet light, the most likely diagnosis is Erythrasma due to the coral-red fluorescence which is characteristic of this condition.

## C. GPT-4V incorrect on easy question

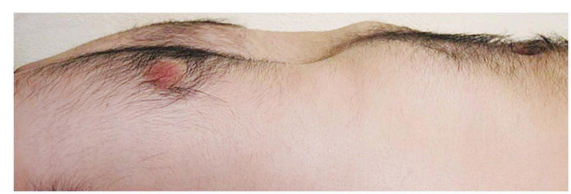

*NEJM* Image Challenge 07-03-2008

**Question**
What is the diagnosis?

**Respondents**
1. Flail chest (4%)
2. Pectus arcuatum (5%)
3. Pectus carinatum (9%)
4. Pectus excavatum (79%) **[max respondents]**
5. Spondylocostal dysostosis (3%)

**GPT-4V (Image Only):** Based on the image provided, the most likely diagnosis is 'Pectus carinatum'.

**GPT-4V (Text Only):** Unfortunately, you haven't provided the case description, so I can't determine the most likely diagnosis from the options given. Please provide the case description so I can assist you further.

**Multimodal GPT-4V:** Based on the provided image and the list of options, the most likely diagnosis is "pectus carinatum."

**Multimodal GPT-4V with CoT:** Alright, let's approach this systematically.
Image Description: The image appears to show the chest region of an individual. There seems to be a prominence or protrusion in the chest wall, specifically at the sternum area.
Possible Diagnoses:
- Flail chest: This is a life-threatening medical condition where a segment of the rib cage breaks and detaches from the rest of the chest wall. The image does not depict multiple broken ribs causing a segment of chest wall to move paradoxically.
- Pectus arcuatum: This is a less commonly referenced condition and is not a standard term in many medical textbooks. It may be a variation or subtype of another condition, but the image does not show any characteristics specifically pointing to this.
- Pectus carinatum: Also known as "pigeon chest", it is characterized by a protrusion of the sternum and ribs. The sternum appears to be pushed outward. This seems to fit with the image provided. Pectus excavatum: Also known as "funnel chest", it is the opposite of Pectus carinatum. In this condition, there is a concave appearance or indentation in the chest wall. This does not fit the image's appearance.
- Spondylocostal dysostosis: This is a rare congenital disorder involving vertebral and rib anomalies. It would likely present with more widespread skeletal anomalies and not just a singular protrusion like seen in the image.

Conclusion: Based on the image and the available options, the most likely diagnosis is Pectus carinatum.

| **D. GPT-4V (Text Only) indeterminate on easy question** | |
|---|---|
| 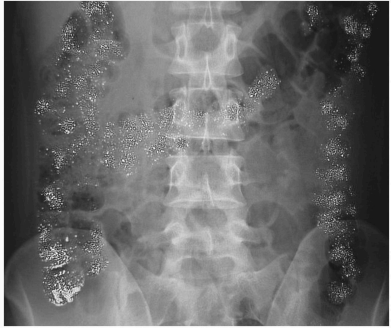 | *NEJM* Image Challenge 01-12-2006<br><br>**Question**<br>What is most likely to account for the findings on this abdominal radiograph?<br><br>**Respondents**<br>1. Schistosomiasis (4%)<br>2. Chronic laxative use (14%)<br>3. Hyperparathyroidism (9%)<br>4. Ischemic colitis (5%)<br>5. Ingestion of a heavy metal (69%) **[max respondents]** |

**GPT-4V (Image Only):** Based on the image provided, the radiological appearance is suggestive of numerous calcifications within the colon, often referred to as "appendicoliths" or "fecaliths" when in the appendix or fecal matter, respectively. The most consistent diagnosis with such a presentation among the options provided would be: Chronic laxative use. Chronic laxative use can lead to calcium or barium salt deposition in the colon, resulting in the characteristic appearance of "milk of calcium" or "soap bubble" appearance on X-rays.

**GPT-4V (Text Only):** To answer this question accurately, I would need to see the abdominal radiograph in question. However, given that I cannot see the image, I can provide some insight into each of the options:\nSchistosomiasis: A parasitic infection that primarily affects the urinary and intestinal systems. Radiographic findings might show calcifications in the bladder wall or intestines. Chronic laxative use: This could result in a radiograph that displays an overall loss of haustral markings in the colon (called "cathartic colon"), which is a result of the long-term stimulant laxative use damaging the nerves and muscles of the colon. Hyperparathyroidism: This condition can result in kidney stones and bone abnormalities like subperiosteal bone resorption. On an abdominal radiograph, you might see nephrocalcinosis or renal stones. Ischemic colitis: This is when blood flow to a part of the colon is reduced. An abdominal radiograph might show bowel wall thickening or thumbprinting due to edema and hemorrhage. Ingestion of a heavy metal: Ingestion of certain heavy metals, like lead or barium, might show up as opaque or dense areas on the radiograph, depending on the metal and its concentration. To determine the most likely cause based on the abdominal radiograph, I'd need a description of the findings or the ability to see the image itself.

**Multimodal GPT-4V:** The image provided is an abdominal radiograph that displays dense stippling throughout the colon. This finding is consistent with ingestion of a radio-opaque substance. Given the options: 'schistosomiasis' is a parasitic infection that affects the liver, bladder, and intestines but does not present as radio-opaque stippling in the colon. 'chronic laxative use' can cause melanosis coli, which presents as a darkening of the colon wall, but it doesn't present as dense stippling. 'hyperparathyroidism' can lead to hypercalcemia and the development of kidney stones, but it does not cause dense stippling in the colon. 'ischemic colitis' results from reduced blood flow to the colon, and it doesn't present with this pattern. The most consistent answer with the findings in the radiograph is: 'ingestion of a heavy metal' such as barium or lead, which can produce the radio-opaque stippling seen throughout the colon. So, the most likely answer based on the findings on this abdominal radiograph is 'ingestion of a heavy metal'.

**Multimodal GPT-4V with CoT:** Certainly, let's analyze the image and the provided options step by step. Image Analysis: The radiograph depicts multiple small opacities throughout the colon. These opacities are uniformly distributed and are consistent in size and appearance. The pattern is reminiscent of a radiopaque substance present within the colon.
Options Analysis:
- Schistosomiasis: This is a parasitic disease caused by blood flukes. While it can affect the gastrointestinal system, it does not produce the type of radiographic appearance shown in the image.
- Chronic laxative use: Chronic use of certain laxatives can result in the accumulation of residual barium, a radiopaque substance. This would appear as multiple small opacities in the colon on an X-ray, similar to what we see in the image.
- Hyperparathyroidism: While this condition can result in various gastrointestinal and bone manifestations, it does not produce the type of radiographic appearance depicted in the image.
- Ischemic colitis: This refers to inflammation and injury of the colon due to reduced blood flow. While ischemic colitis can cause changes in the appearance of the colon on imaging, it doesn't typically manifest as multiple small opacities.
- Ingestion of a heavy metal: Ingested heavy metals, such as lead or bismuth, can appear as radiopaque substances in the gastrointestinal system. However, the pattern of distribution and the consistency of the opacities seen here are more suggestive of residual barium from laxatives rather than ingested heavy metals.

Conclusion: Based on the radiographic appearance and the provided options, the most likely diagnosis accounting for the findings on this abdominal radiograph is 'Chronic laxative use'. This pattern is suggestive of the retention of barium, a radiopaque substance used in some laxatives.

**Figure 3. Multimodal Reasoning Explanations Provided by GPT-4V.** Four example cases and GPT-4V output for GPT-4V (Image Only), GPT-4V (Text Only), Multimodal GPT-4V, and Multimodal GPT-4V with CoT (chain of thought) are shown. Multimodal CoT (chain of thought) uses the same prompt but adds "Let's think step by step" to the end of the question to elicit reasoning. Correct answers are highlighted in green and the respondent distribution is also shown. We ranked cases by human performance and selected **A-B.** Difficult examples that illustrate GPT-4V reasoning multimodally and from text alone, as well as easy examples where **C.** GPT-4V answers incorrectly and **D.** GPT-4V (Text Only) refuses to answer without the image.

# Figure 4. Multimodal Evaluation on Clinicopathological Conferences

**A. Accuracy of Identifying the Correct Diagnosis in the Differential for GPT-4V and GPT-4V Turbo Given Either Text Only or Text and Images from the Presentation of Case**

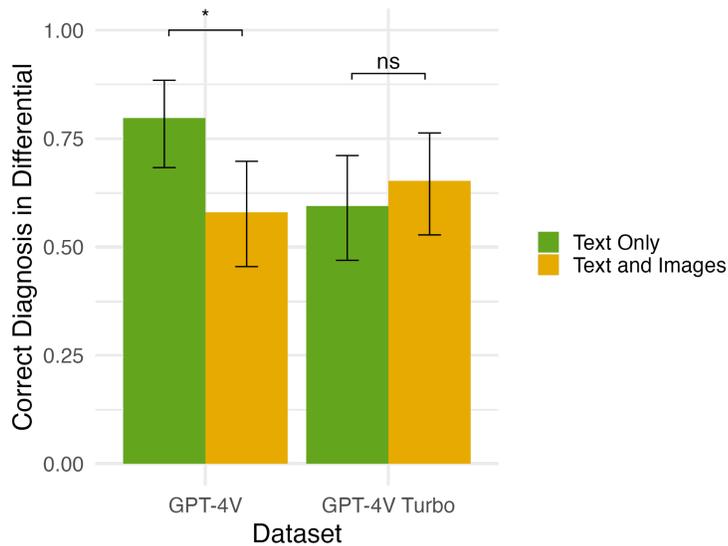

**B. Quality of GPT-4V Turbo Differential Diagnosis With Different Modalities from CPCs**

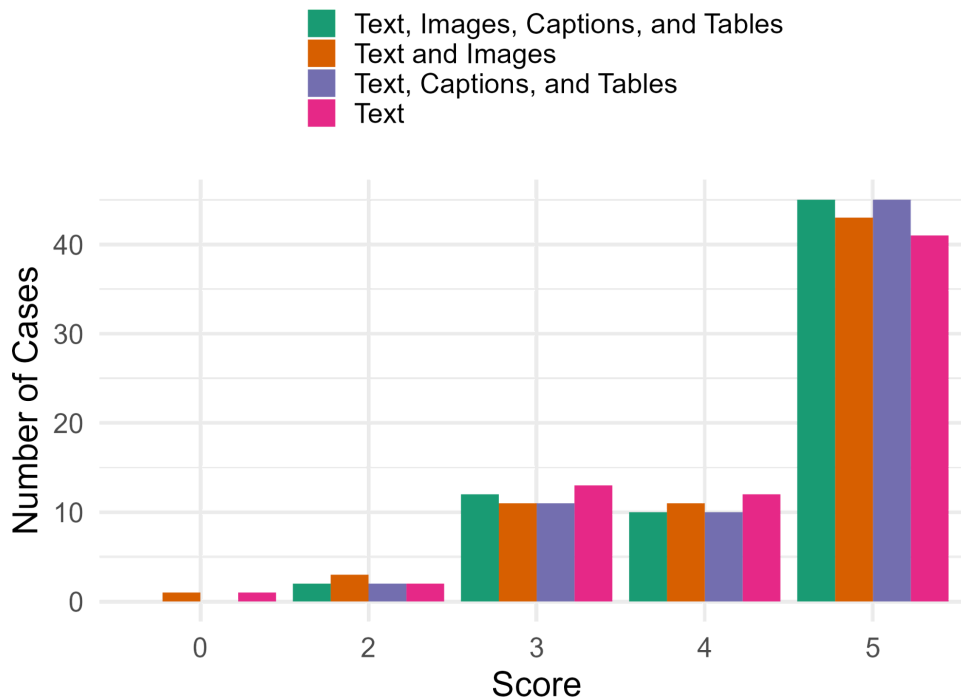

**Figure 4. Multimodal Evaluation on Clinicopathological Conferences**.
**A.** Comparing the accuracy of GPT-4 and GPT-4V Turbo with and without images in predicting the correct diagnosis within the differential. *: $p < 0.05$, **: $p < 0.01$, ***: $p < 0.001$, ns: not significant. **B.** Physician-scored performance (0 = worst, 5 = best) of Multimodal GPT-4V Turbo on 69 *NEJM* clinicopathological conferences (CPCs) in the "Presentation of Case" section of the CPCs. The models were evaluated on different combinations of (a) the text alone, (b) the text and images, (c) the text images, captions, and tables, and (d) the text, captions, and tables. Tables were included as a textual comma separated value (CSV) format and images were attached in their full resolution. Each case was scored by a board-certified physician using the Bond et al. *Gen Intern Med*. 2012 criteria.